\documentclass{article}

\usepackage{graphicx}
\usepackage{subcaption}
\usepackage{amsmath,amssymb}
\usepackage{booktabs}
\usepackage{tabularx}

\usepackage{hyperref}
\usepackage{xurl}       
\usepackage[
  backend=biber,
  style=numeric,
  doi=false,
  url=false,
  isbn=false
]{biblatex}
\title{Emergent Coordination and Phase Structure \\
in Independent Multi-Agent Reinforcement Learning}

\author{%
  Azusa Yamaguchi \\
  School of Physics and Astronomy \\
  University of Edinburgh\\
  \texttt{ayamaguc@staffmail.ed.ac.uk} \\
  }

\begin{document}

\maketitle

\begin{abstract}
A clearer understanding of when coordination emerges, fluctuates, or collapses
in decentralized multi-agent reinforcement learning (MARL) is increasingly sought in order to characterize the dynamics of multi-agent learning systems.
We revisit fully independent Q-learning (IQL) as a minimal decentralized testbed 
and run large-scale experiments across environment size $L$ and agent density 
$\rho$.  
We construct a phase map using two axes---cooperative success rate (CSR) and a 
stability index derived from TD-error variance---revealing three distinct regimes: 
a coordinated and stable phase, a fragile transition region, and a jammed/disordered 
phase.  
A sharp \emph{double} Instability Ridge separates these regimes and corresponds to 
persistent kernel drift, the time-varying shift of each agent's effective transition 
kernel induced by others' policy updates.  
Synchronization analysis further shows that temporal alignment is required for 
sustained cooperation, and that drift–synchronization competition generates 
the fragile regime.  
Ablation without agent identifiers removes drift entirely and collapses the 
three-phase structure, demonstrating that small inter-agent asymmetries are a 
necessary driver of drift.  
Overall, our results show that decentralized MARL exhibits a coherent phase 
structure governed by the interaction between scale, density, and kernel drift, 
suggesting that emergent coordination behaves as a distribution--interaction 
driven phase phenomenon.
\end{abstract}

\section{Introduction}

A central aim of artificial intelligence is to develop agents that can perceive,
adapt, and interact with others in complex environments. Reinforcement learning
(RL) provides a basic framework for learning through experience, with many of its
successes demonstrated in single-agent domains
\cite{Arulkumaran2017DeepRL,Li2017DeepRL}.

In multi-agent reinforcement learning (MARL), however, several agents update
their policies concurrently, altering each other's transition dynamics and
creating an inherently non-stationary learning problem \cite{hernandezleal2019}.
Such non-stationarity is frequently associated with instability, divergence, and
failures of coordination.

A number of influential approaches mitigate these challenges by introducing
centralized critics or structural biases. CTDE methods such as MADDPG
\cite{lowe2017multi} and COMA \cite{foerster2018coma} stabilize learning through
centralized information, while value-decomposition approaches such as VDN
\cite{sunehag2017vdn} and QMIX \cite{rashid2018qmix} impose additive or
monotonic structures to promote cooperation. These techniques are highly
effective but rely on explicit architectural assumptions that constrain the joint
value landscape \cite{mahajan2020,rashid2020weightedqm}. Consequently, they offer
limited insight into how coordination might emerge---or collapse---\emph{without}
centralized bias.

Emergent coordination has also been examined from perspectives including
evolutionary games \cite{Tuyls2005,Bloembergen2015}, social dilemmas
\cite{leibo2017}, and collective multi-agent systems
\cite{Hindes2021,Bratsun2024,rubenstein2014}. These studies suggest that MARL
behaviors can resemble physical collective phenomena. 
However, because coordination is actively enforced in these frameworks, they offer
limited insight into how coordination might arise—or fail spontaneously—in fully
decentralized settings \emph{without centralized critics, structural assumptions,
or explicit coordination mechanisms}.

\medskip
In this work, we revisit Independent Q-Learning (IQL), which provides a clean
testbed for examining emergent coordination because it operates \emph{without
imposing centralized inductive biases or decompositional structures}.


Although IQL is often regarded as unstable or limited
\cite{tan1993multi,foerster2018}, it is precisely its lack of centralized
critics, structural decomposition, or enforced cooperation that makes it an
ideal setting for studying spontaneous coordination. By scanning a broad range
of $(L,\rho)$ conditions, we show that this minimal MARL system exhibits a
rich phase structure driven by distributional non-stationarity.

\medskip
\textbf{Our contributions are as follows:}

\begin{enumerate}
    \item \textbf{A phase map of coordination and stability.}
    Combining the coordination success rate (CSR) with a TD-error--variance
    stability index $S$, we identify coordinated, fragile, and jammed/disordered
    regimes separated by a \emph{double} instability ridge in the $(L,\rho)$
    plane.

    \item \textbf{Kernel drift as a mechanism for MARL non-stationarity.}
    Temporal analysis of TD-error variance and gradient-norm variance shows that
    instability correlates strongly with kernel drift—the time-varying drift of
    the effective transition kernel induced by others' policy updates.

    \item \textbf{Synchronization as a requirement for sustained coordination.}
    Arrival-time spread and co-reach statistics reveal that temporal
    synchronization is necessary for maintaining coordination and that
    insufficient synchronization characterizes the fragile regime.

    \item \textbf{Spontaneous coordination in decentralized MARL.}
    Even without centralized bias, certain scale–density combinations yield
    stable coordinated behavior, suggesting that decentralized MARL can exhibit
    phase-transition-like phenomena.
\end{enumerate}

Together, these results provide a unified perspective in which emergent
coordination, fluctuation, and collapse arise from interactions among scale,
density, and kernel drift, forming a coherent phase structure.

\section{Methods}

This section describes the environment, learning setup, and evaluation metrics used to analyze
emergent coordination, non-stationarity, and kernel drift in fully decentralized Independent
Q-Learning (IQL).

\subsection{Environment: Grid-Based Multi-Agent Navigation}

We use an $L\times L$ grid world with $L\in\{8,16,24,32\}$, containing $N$ agents and a single
goal placed without overlap.  
Agent density is defined as
\[
\rho_{\mathrm{agents}} = N/L^2,\qquad 
\rho\in\{0.03125,\,0.0625,\,0.125,\,0.25,\,0.5\},
\]
excluding $(L=32,\rho=0.5)$ due to computational cost.

Agents choose from
\[
A=\{\texttt{stay},\texttt{up},\texttt{down},\texttt{left},\texttt{right}\},
\]
and receive $-0.005$ per step, $+1$ upon reaching the goal, and a \texttt{hold} action thereafter
which no longer affects the transition kernel.

Episodes terminate once the accumulated reward reaches the target score
$0.8N$, or when the maximum horizon of $8L$ steps is reached.
Each condition is trained for up to 1500 episodes.
A single goal ensures that changes in $(L,\rho)$ naturally induce sparsity and congestion.

\subsection{Learning Algorithm: Parameter-Shared Double DQN}

All agents share parameters and learn via Double~DQN.  
Given online parameters $\theta$ and target parameters $\theta^{-}$, the TD target is
\[
a'=\arg\max_a Q_\theta(\tilde{s},a),\qquad
y=r+\gamma(1-d)\,Q_{\theta^-}(\tilde{s},a').
\]

We optimize the Huber loss
\[
L(\theta)=\mathbb{E}[\ell_{\mathrm{Huber}}(y-Q_\theta(s,a))],
\]
and update the target network via Polyak averaging:
\[
\theta^{-}\leftarrow(1-\tau)\theta^{-}+\tau\theta.
\]

Exploration follows exponentially decaying $\epsilon$-greedy, and evaluation uses $\epsilon=0$.
Parameter sharing improves sample efficiency while retaining decentralized execution.

\subsection{Network, Optimization, and Replay Buffer}

We use a two-layer MLP (128 units each, ReLU), gradient clipping, and the Adam optimizer
(PyTorch defaults $\beta_{1}=0.9,\beta_{2}=0.999$).  
Full hyperparameters appear in Appendix~\ref{tab:hyper_param}.

A shared replay buffer $\mathcal{D}$ stores $10^5$ transitions;
training updates start once it contains at least $1{,}500$ transitions,
with a mini-batch size of 64.
To permit spontaneous role differentiation despite parameter sharing, each transition augments
observations with a one-hot agent identifier:
\[
o^{(i)}_{\mathrm{ID}} = [\,o^{(i)} \,\|\, \mathrm{ID}_i\, ].
\]

\subsection{Phase Structure: Coordination and Drift Metrics}

We characterize each $(L,\rho)$ condition using two axes:
\begin{itemize}
    \item \textbf{Cooperative Success Rate (CSR)},  
    \item \textbf{Stability Index $S$} based on TD-error variance.
\end{itemize}

\subsubsection{Cooperative Success Rate (CSR)}

During evaluation episodes $K_{\mathrm{eval}}$ (with $\epsilon=0$),
\[
\mathrm{CSR}(L,\rho)=
\frac{1}{K_{\mathrm{eval}}}\sum_{k=1}^{K_{\mathrm{eval}}}
\mathbf{1}\{\text{all agents reached}\}.
\]

\subsubsection{Stability Index $S$: TD-Error Variance as Kernel Drift Proxy}

Non-stationarity arises from temporal drift in the effective transition kernel $P_i^t$ caused by
policy updates of other agents.  
We estimate this effect using the per-episode variance of the TD error
\[
\delta_t = y_t - Q(s_t,a_t),
\]
denoted $v(L,\rho)$.  
Normalizing by the maximum variance across all conditions,
\[
S(L,\rho)= 1 - \frac{v(L,\rho)}{v_{\max}},
\]
where high $S$ indicates weak drift (stable) and low $S$ indicates strong drift (unstable).
Gradient-norm variance is analyzed in the Appendix.

\subsubsection{Reference Point and Phase Distance}

Thresholds are defined as the 60th percentiles of CSR and $S$:
\[
\tau_{\mathrm{CSR}}=\mathrm{Perc}_{60}(\mathrm{CSR}),\qquad
\tau_S=\mathrm{Perc}_{60}(S).
\]

The phase distance is
\begin{equation}\label{eq:distance}
d_{\mathrm{phase}}(L,\rho)=
\sqrt{(\mathrm{CSR}-\tau_{\mathrm{CSR}})^2 + (S-\tau_S)^2}.
\end{equation}

This yields three regimes:
\begin{itemize}
    \item \textbf{Stable Coordinated Phase}: high CSR, high $S$; kernel drift suppressed.
    \item \textbf{Fragile Transitional Region}: near the minimum of $d_{\mathrm{phase}}$; coordination fluctuates.
    \item \textbf{Jammed / Disordered Phase}: low CSR, low $S$; congestion-induced stagnation.
\end{itemize}

\subsection{Experimental Setup}

\subsubsection{Training}

\begin{itemize}
    \item 1{,}500 episodes,
    \item Maximum steps $= 8L$,
    \item 50 random seeds (mean and 95\% confidence interval),
    \item Learning rate $1.5\times10^{-4}$, discount $\gamma=0.95$, Polyak $\tau=10^{-3}$.
\end{itemize}

\subsubsection{Evaluation Conditions}

CSR is computed during a separate evaluation phase using greedy policies ($\epsilon=0$).
In contrast, the stability index $S$ is derived from TD-error statistics recorded \emph{during training},
under the standard $\epsilon$-greedy exploration schedule. 
These two quantities are then combined to construct the phase maps.

\subsubsection{Compute Resources}

Experiments were conducted on the University of Edinburgh’s \textit{Eddie} HPC cluster  
(Rocky Linux~9, Altair GridEngine).

\section{Results}

\subsection{Global Structure of Coordination and Non-coordination: CSR--Stability Phase Map}

\begin{figure*}[t]
\centering
\includegraphics[width=0.95\textwidth]{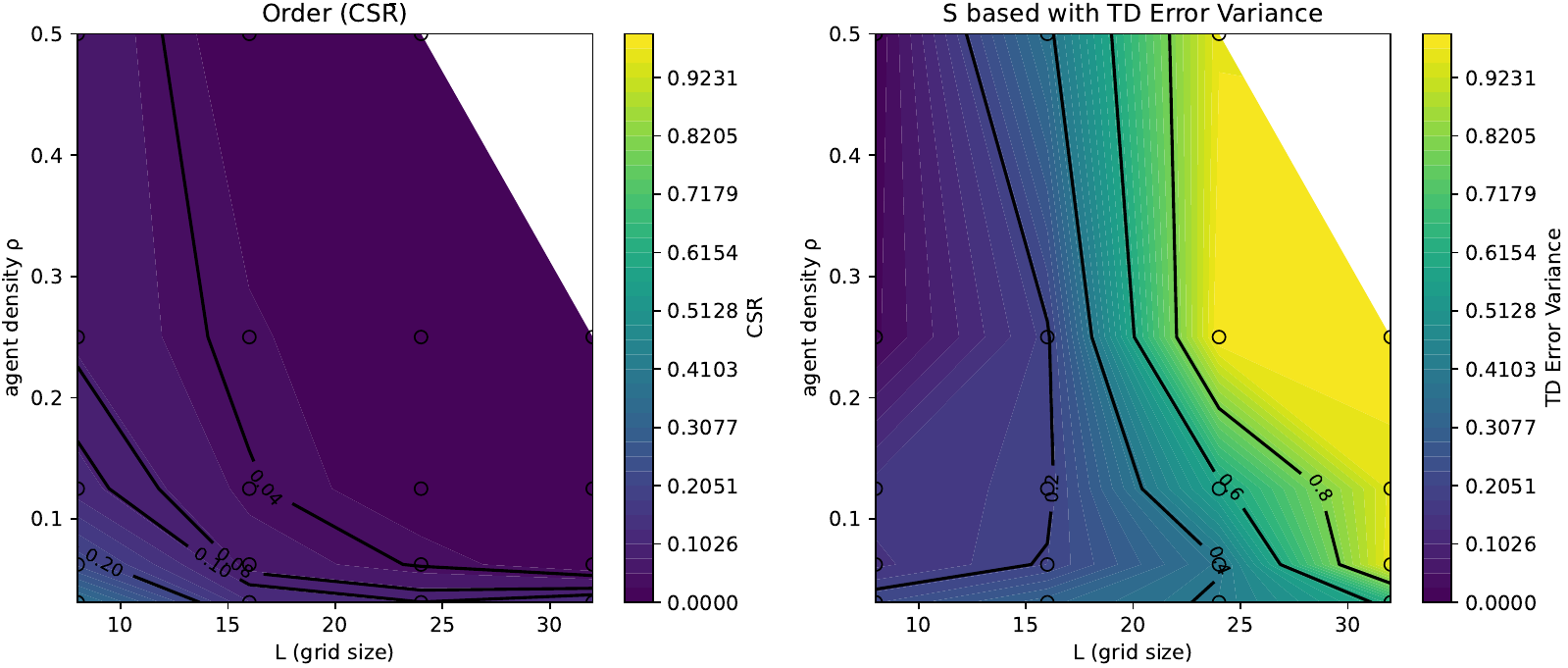}
\caption{
Cooperative success rate (CSR, left) and stability index $S$ (right) 
for all $(L,\rho)$ conditions, computed from the last 25\% of episodes.
Coordination and stability occur only at small scales and low densities,
while both metrics collapse sharply as scale or density increases.
The low-$S$ region marks strong non-stationarity and forms an Instability Ridge.
}
\label{fig:order}
\end{figure*}

Fig.~\ref{fig:order} shows CSR and the stability index $S$ (TD-error–variance-based)
for 19 conditions ($L\in\{8,16,24,32\}$ and $\rho\in\{0.03125,\dots,0.5\}$),
computed from the last 25\% of training episodes.

At small scales and low densities, CSR and $S$ are simultaneously high, indicating that coordinated behavior can emerge even under fully decentralized IQL.

As density increases, CSR drops sharply toward zero across all scales, while TD-error variance increases.  
This pattern suggests that congestion-induced exploration difficulty amplifies kernel drift, making the policy updates increasingly unstable.

At fixed density, increasing scale also degrades both CSR and $S$ monotonically, revealing that coordination becomes progressively fragile as exploration cost and agent interactions intensify.  
Taken together, the CSR–$S$ axes provide a clear separation between coordinated, partially coordinated, and non-coordinated regimes, and illustrate that coordination in independent MARL is strongly dependent on environmental scale and density, collapsing rapidly with growing non-stationarity.

\subsection{Phase Geometry via Distance from the Coordinated Attractor}

\begin{figure*}[t]
\centering
\includegraphics[width=0.95\textwidth]{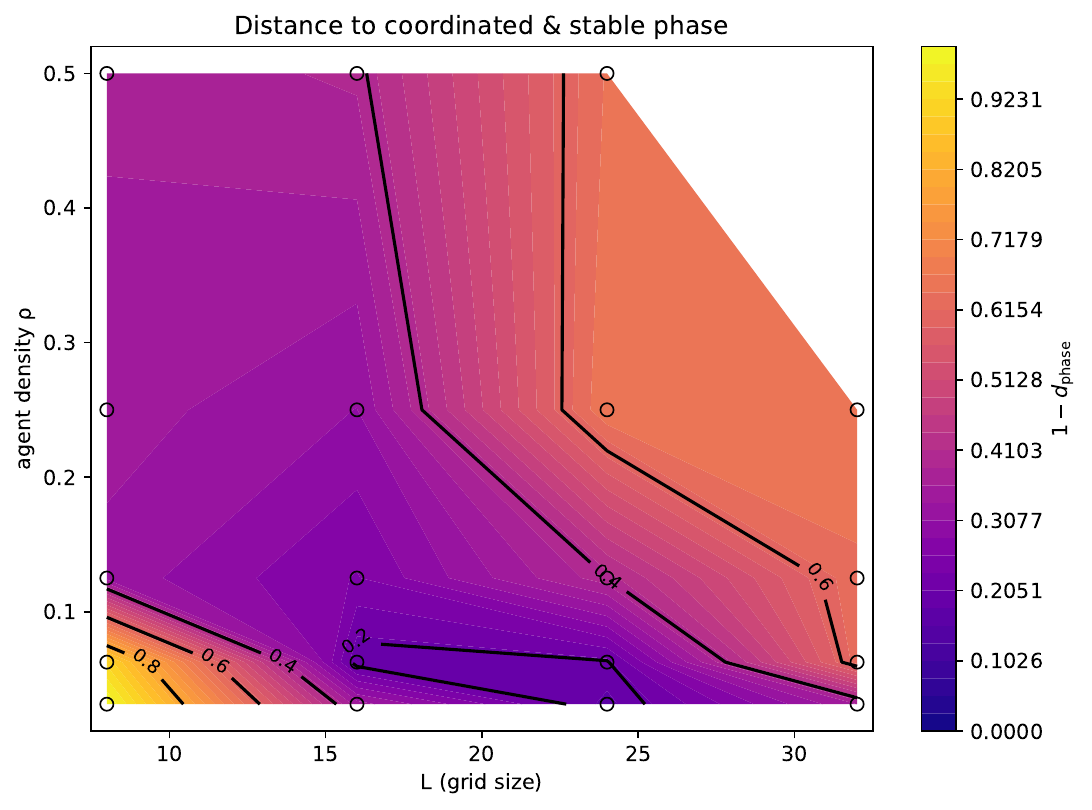}
\caption{
Phase geometry based on the normalized distance $d_{\mathrm{phase}}$.
Two contour lines near $d_{\mathrm{phase}}\!\approx\!0.4$ form a double Instability Ridge.
The low-$L$, low-$\rho$ region corresponds to coordinated and stable behavior, 
the region between the ridges to a fragile transitional regime,
and larger scales to jammed/disordered outcomes.
}
\label{fig:distance}
\end{figure*}

Fig.~\ref{fig:distance} visualizes the normalized distance $d_{\mathrm{phase}}$ from Eq.~\ref{eq:distance}.
A notable feature is the presence of two contour lines around $d_{\mathrm{phase}}\approx 0.4$,
forming a double Instability Ridge that separates coordinated and non-coordinated behavior.

This structure partitions the $(L,\rho)$ space into three regimes:

\begin{itemize}
\item \textbf{Coordinated \& Stable Phase} ($d_{\mathrm{phase}}> 0.4$ in the low-$L$, low-$\rho$ region ):  
High CSR and high $S$, with weak drift and stable synchronization.  
Convergence toward a coordinated attractor is consistently observed.

\item \textbf{Fragile Region} ($d_{\text{phase}} < 0.4$,between the two ridges):  
Synchronization and collapse alternate, producing non-monotonic and transitional dynamics.  
Kernel drift competes with synchronization, yielding only temporary or partial coordination.

\item \textbf{Jammed/Disordered Phase} ($d_{\mathrm{phase}}>0.4$ at larger $L$):  
CSR is near zero and $S$ remains low.  
Although kernel drift weakens, gradient noise later dominates, leading to irreversible jammed/disordered behavior.
\end{itemize}

This three-part geometry indicates that IQL exhibits bifurcating dynamics between a coordinated attractor and drift-driven instability, resembling a phase transition.

\subsection{Synchronization and Partial Coordination Near the Instability Ridge}

\begin{figure*}[ht]
\centering
\begin{subfigure}{0.48\textwidth}
\includegraphics[width=\linewidth]{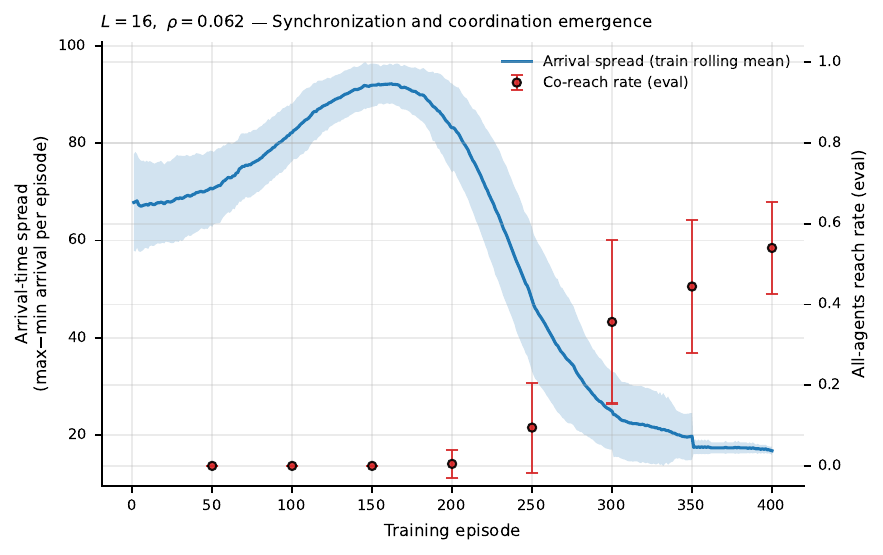}
\caption{$L=16,\rho=0.0625$ (pre-ridge)}
\end{subfigure}
\hfill
\begin{subfigure}{0.48\textwidth}
\includegraphics[width=\linewidth]{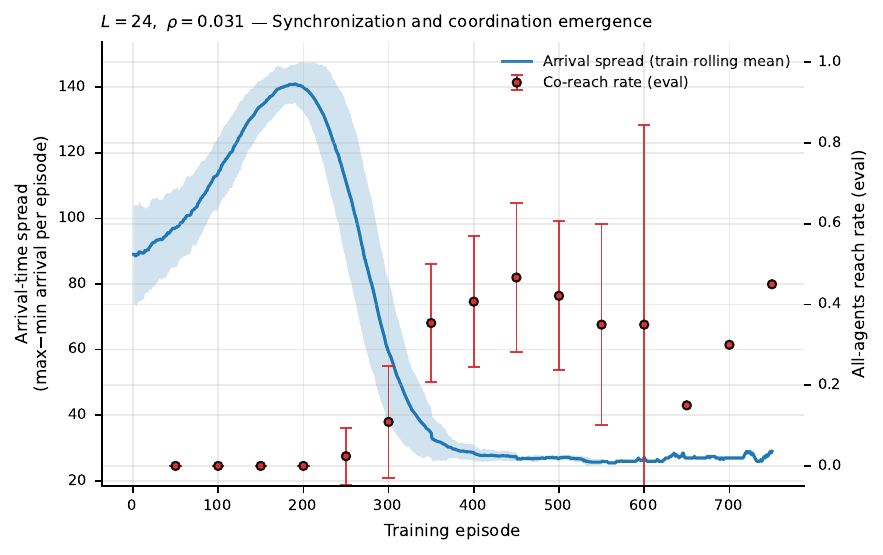}
\caption{$L=24,\rho=0.03125$ (on the ridge)}
\end{subfigure}
\caption{Temporal profiles of arrival-time spread (synchronization) and co-reach rate 
for two representative conditions near the Instability Ridge.
Pre-ridge ($L{=}16,\rho{=}0.0625$) shows rapid convergence toward coordination,
whereas on-ridge ($L{=}24,\rho{=}0.03125$) exhibits alternating collapse--recovery cycles.}
\label{fig:coord_arrive}
\end{figure*}

Fig.~\ref{fig:coord_arrive} compares synchronization (arrival-time spread) and coordination (co-reach) dynamics for two ridge-adjacent conditions.

In the pre-ridge case, drift remains weak, spread collapses quickly, and co-reach increases smoothly—reflecting stable convergence to a coordinated attractor.

On the ridge, spread and co-reach repeatedly undergo collapse–recovery cycles.  
This oscillatory pattern reflects competition between synchronization and kernel drift, giving rise to fragile coordination that does not persist.

These observations support the view that synchronization is a necessary condition for coordination, and that drift near the ridge disrupts this mechanism.
(High-density cases transition into jammed or oscillatory behavior; see Appendix Fig.~\ref{fig:synccoord_2x2}.)

\subsection{Kernel Drift vs.~Gradient Noise: Two Forms of Instability}

\begin{figure*}[t]
\centering
\includegraphics[width=0.95\textwidth]{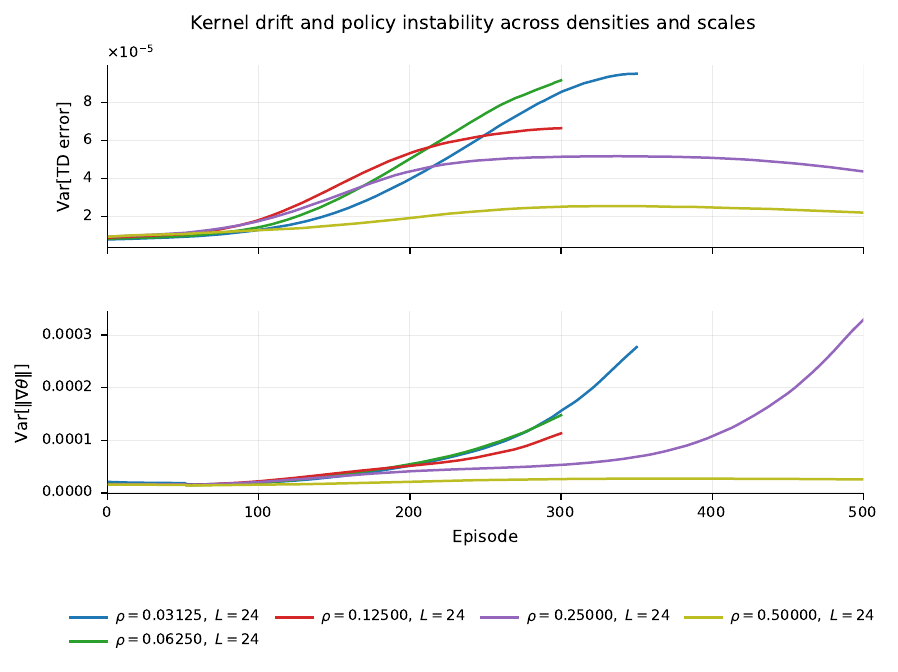}
\caption{
TD-error variance (top) and gradient-norm variance (bottom) for $L=24$ across densities.
The dominant source of instability differs by density: 
persistent variance growth near the Ridge, 
oscillatory behavior on the Ridge, 
early saturation outside the Ridge, 
and late-stage divergence at high densities.These four densities were selected because they represent pre-ridge, on-ridge, post-ridge, and high-density regimes, respectively.
}
\label{fig:L24_kd}
\end{figure*}

Fig.~\ref{fig:L24_kd} shows that the dominant source of instability changes with density:

\begin{itemize}
\item \textbf{Near-ridge ($\rho=0.0625$)}:  
TD variance grows throughout training, and gradient variance increases later.  
Drift does not settle, preventing convergence to a coordinated fixed point.

\item \textbf{On-ridge ($\rho=0.03125$)}:  
TD variance remains high, and gradient variance increases gradually, producing fragile oscillatory dynamics.

\item \textbf{Outside-ridge ($\rho=0.125$)}:  
Both variances saturate early, reflecting premature lock-in to incomplete patterns rather than coordination.

\item \textbf{High densities ($\rho\ge 0.25$)}:  
Drift weakens but gradient variance diverges in late training, causing loss of update coherence and jammed/disordered outcomes.
\end{itemize}

These trends indicate that the Instability Ridge corresponds to a transition 
from drift-dominated to gradient-noise–dominated instability, offering a unified perspective on coordination, partial coordination, and collapse.  
Complete time-series statistics of TD error, gradient norms, and their variances across all conditions are provided in Appendix Fig.~\ref{fig:learning_stability_withID}, 
with further discussion in Appendix~\ref{seq:Learning_stability_WITHID}.

\subsection{Macroscopic Statistics of the Coordination Transition: Synchronization and Density Thresholds}

\begin{figure*}[t]
\centering
\begin{subfigure}{0.48\textwidth}
\includegraphics[width=\linewidth]{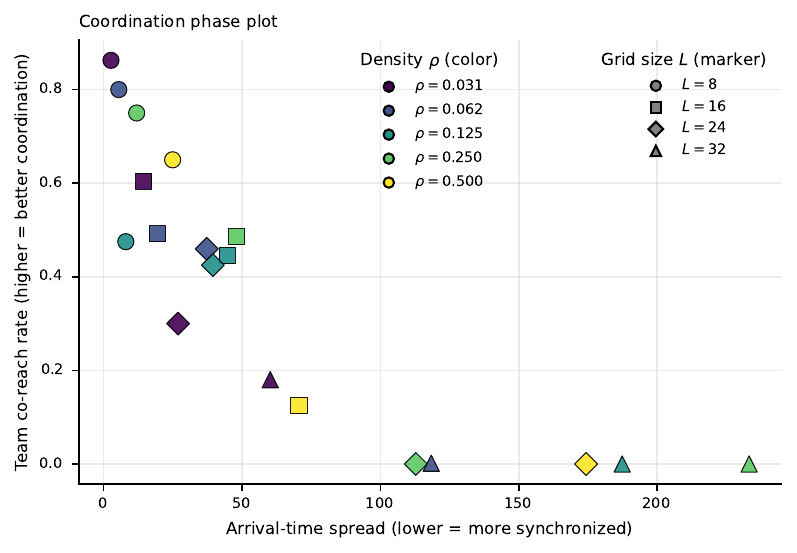}
\caption{}\label{fig:sync_coord}
\end{subfigure}
\hfill
\begin{subfigure}{0.48\textwidth}
\includegraphics[width=\linewidth]{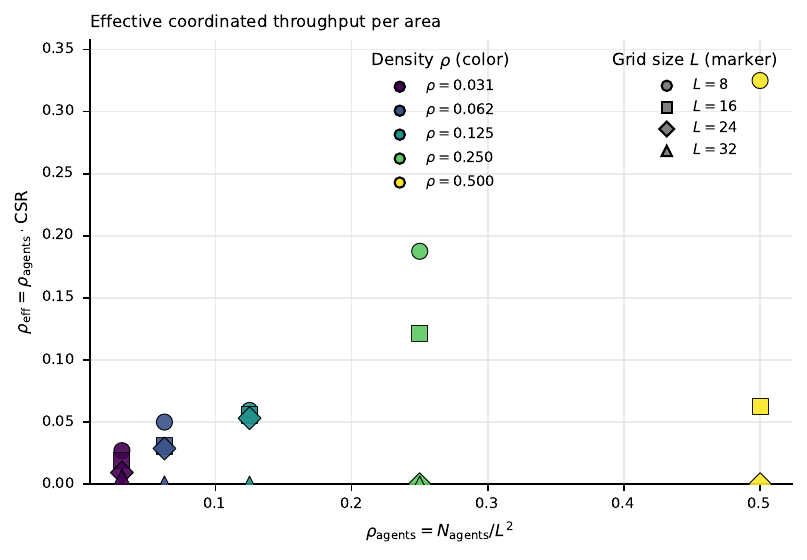}
\caption{}\label{fig:rho_eff}
\end{subfigure}
\caption{(Left) Relationship between arrival-time spread and co-reach across all conditions. 
High co-reach occurs only when spread is small; 
large fluctuations appear near the Instability Ridge.
(Right) Effective coordinated throughput 
$\rho_{\mathrm{eff}}=\rho_{\mathrm{agents}}\mathrm{CSR}$,
showing scale-dependent critical densities where throughput peaks and then declines.}
\label{fig:sync_coord_rhoeff}
\end{figure*}

Fig.~\ref{fig:sync_coord} shows that only conditions with small spread achieve high co-reach.
The absence of points with large spread and high co-reach indicates that non-synchronized coordination is not observed.
Large fluctuations near the ridge again reflect drift--synchronization competition.

Fig.~\ref{fig:rho_eff} shows the effective coordinated throughput
$\rho_{\mathrm{eff}} = \rho_{\mathrm{agents}} \cdot \mathrm{CSR}$,
which reveals scale-dependent critical densities $\rho_{\mathrm{crit}}(L)$.
At medium and large scales, throughput peaks at intermediate densities and drops sharply at high densities.
This behavior mirrors the synchronization and coordination collapse observed in Fig.~\ref{fig:coord_arrive}, appearing as a macroscopic phase transition.

\section{Discussion}

These findings position kernel drift not merely as a correlate of instability, but as a mechanistic driver whose growth rate predicts the onset of fragile and jammed regimes.

Non-stationarity is a central challenge in multi-agent reinforcement learning (MARL),
arising from the fact that each agent's policy update modifies the effective transition
dynamics faced by all others. Prior work has examined how replay-induced drift
\cite{foerster2018}, non-convex--non-concave game structures \cite{zhang2021},
and regularization or learning-rate control \cite{papoudakis2019}
contribute to instability.
This interpretation complements prior taxonomies of non-stationarity Hernandez-Leal et al. (2019), but identifies kernel drift as a unifying mechanism driving the observed phase structure.

Our results suggest that these diverse forms of non-stationarity can be viewed through a
single unifying mechanism: \emph{kernel drift}, the distributional drift in each agent’s
effective transition kernel. For agent $i$, the kernel
\begin{equation}\label{Eq:KL}
P_i^t(s'|s,a_i)=\sum_{a_{-i}}P(s'|s,a_i,a_{-i})
\prod_{j\ne i}\pi_j^{t}(a_j|s)
\end{equation}
fluctuates as other agents update, producing a drift term $\Delta P_i^t$.
These fluctuations amplify the variance of TD targets,
create a mismatch between replayed and current dynamics,
and push policy updates away from stable fixed points.
This mechanism is qualitatively consistent with observations in
mean-field MARL, replicator dynamics, and actor–critic oscillations.

The resulting phase diagram reveals three regimes shaped by the interaction between
kernel drift and gradient noise:
\begin{itemize}
\item \textbf{Stable coordinated phase:} At low density and small scale,
kernel drift is weak, synchronization holds, and learning consistently
converges toward a coordinated attractor.
\item \textbf{Fragile transitional phase:} Near the Instability Ridge,
kernel drift grows critically and competes with synchronization,
producing alternating collapse–recovery cycles and highly non-monotonic dynamics.
\item \textbf{Failure phase (jammed/disordered):} Outside the ridge and at
higher densities, kernel drift weakens but gradient noise becomes dominant,
leading to early lock-in or jammed/disordered behavior from which
coordination does not recover.
\end{itemize}

Notably, with a single shared goal, increasing the scale $L$ reduces collision-induced
asymmetries, weakening kernel-drift sources and causing the dynamics at $L=24$ and $L=32$
to converge toward similar regimes—an observation that further motivates the
ID-removal ablation examined below.

The ablation in Appendix~\ref{seq:app_noid} further clarifies the mechanism:
removing the agent ID restores full input symmetry, yielding identical
transition kernels for all agents, $\Delta P_{\mathrm{brk}}^t\equiv 0$.
Kernel drift is therefore canceled over time.
TD-error and gradient-norm variances remain low and flat,
CSR becomes density-independent,
and neither coordinated nor transitional phases emerge.
This provides direct evidence that the observed phase structure is not
an artifact of algorithmic bias, but rather arises from small symmetry-breaking
differences that accumulate through distributional interaction.

These observations align with prior work on role emergence or symmetry breaking
in MARL \cite{baker2020}, evolutionary games \cite{Tuyls2005,Bloembergen2015},
and collective robotics \cite{Bratsun2024}.
The phase-transition–like structure observed here appears to be a spontaneous
phenomenon rooted in distributional interaction rather than an explicit
coordination mechanism.

Overall, kernel-drift dynamics highlight how MARL learning trajectories may be
understood as collective processes shaped by distribution-level feedback. A
rigorous theoretical account remains open, but our empirical findings point
toward promising connections with distributional or mean-field stability analyses.
\section{Conclusion}

This work examined how independent multi-agent reinforcement learning (IQL),
despite having no centralized critic or structural coordination bias,
exhibits systematic patterns of emergent coordination, fragility, and failure
across environment scale $L$ and agent density $\rho$.
By combining cooperative success (CSR) with a stability index based on TD-error
variance, we constructed phase diagrams that reveal three regimes:
\begin{itemize}
\item \textbf{Coordinated \& stable:} synchronization holds and learning converges
toward a coordinated attractor.
\item \textbf{Fragile coordination:} kernel drift competes with synchronization,
producing collapse–recovery oscillations.
\item \textbf{Failure / jammed--disordered:} kernel drift weakens but gradient noise
dominates, preventing recovery of coordination.
\end{itemize}

The Instability Ridge emerges as a sharp transition boundary where kernel drift
amplifies and the learning dynamics switch from drift-dominated to
gradient-noise–dominated behavior. These observations indicate that MARL learning
trajectories may exhibit phase-transition–like structure shaped by distributional
interaction.

Ablation experiments further showed that removing agent IDs eliminates the entire
phase structure: TD-error variance stays uniformly low, CSR becomes flat across
densities, and neither coordination nor fragility nor collapse emerges. This
supports the interpretation that the observed dynamics arise from spontaneous
symmetry breaking and distributional interaction, rather than from explicit
coordination mechanisms.

Taken together, our results suggest that coordination in MARL may be understood
as a spontaneous, distribution-driven phenomenon governed by physical-style
parameters such as scale, density, kernel drift, and gradient noise. Treating
kernel drift as a distributional fluctuation in effective transition dynamics
provides a unifying view on non-stationarity and offers a foundation for future
stability analyses based on distributional or mean-field models.

\subsection*{Implications for Multi-Agent Economics and Financial Markets}

The structure observed here—emergence, fluctuation, and breakdown of coordination
driven by distributional interactions—has natural parallels in multi-agent
economics and market microstructure. Many-agent systems in which actions modify
the distributional environment that, in turn, shapes optimal behavior share the
same feedback structure as MARL.

The phenomena observed in this work—
emergence of coordination (price stability), transient or partial synchronization,
critical-density collapse, and drift-induced deviation from equilibrium—
suggest that scale, density, and distributional drift may provide useful
explanatory principles for when markets stabilize and when they destabilize.

These connections point toward future work on the stability of many-agent
financial models, critical phenomena in order-book dynamics, and MARL-based
market simulations. Understanding coordination and its breakdown through the
lens of phase geometry may offer a promising direction for integrating ideas
from AI, economics, and statistical physics.

\section{Acknowledgments}
A.Y. received support from a sponsored research agreement with Intel Corporation, which provided partial funding for this research.

\printbibliography

\appendix
\section{Appendix}

This appendix provides additional analyses and ablation studies that complement the main text.
In particular, we present:
(i) full hyperparameter specifications,
(ii) extended phase maps,
(iii) additional examples of synchronization--coordination dynamics,
(iv) scale-dependent properties of kernel-drift indicators, and
(v) ablations in which agent identifiers are removed to examine the collapse of the phase structure.

\subsection{Hyperparameter Details}

\begin{table}[h]
\centering
\small
\caption{Hyperparameters used in all experiments.}\label{tab:hyper_param}
\begin{tabularx}{\linewidth}{@{} l l X @{}}
\toprule
Item & Symbol & Value / Setting \\
\midrule
Number of episodes (max) & -- & $1500$ \\
Max steps per episode & $T_{\max}$ & $8L$ \\
Number of random seeds per condition & -- & $50$ \\
\midrule
Learning rate & $\alpha$ & $1.5 \times 10^{-4}$ \\
Discount factor & $\gamma$ & $0.95$ \\
Polyak coefficient (soft target update) & $\tau$ & $10^{-3}$ \\
Optimizer & -- & Adam ($\beta_1=0.9,\ \beta_2=0.999$)\textsuperscript{*} \\
Batch size & $B$ & $64$ \\
Replay buffer capacity & $|\mathcal{D}|$ & $10^{5}$ \\
Warm-up transitions & -- & $1500$ \\
\midrule
Initial exploration rate & $\epsilon_{\mathrm{start}}$ & $1.0$ \\
Minimum exploration rate & $\epsilon_{\min}$ & $0.01$ \\
Exploration decay (per episode, exponential) & $\lambda$ & $0.98$ \\
Exploration during evaluation & $\epsilon_{\mathrm{eval}}$ & $0$ (greedy) \\
\midrule
Network architecture & -- & Input–FC(128, ReLU)–FC(128, ReLU)–Output \\
Activation function & -- & ReLU \\
Gradient clipping & -- & $\|\nabla\|_2 \le 1.0$ \\
Loss function & -- & Huber loss (smooth L1) \\
\bottomrule
\end{tabularx}

\vspace{0.5em}
\footnotesize{\textsuperscript{*}PyTorch's default hyperparameters for Adam were used.}

\end{table}

\subsection{Stability Index Based on Gradient-Norm Variance and the Corresponding Phase Maps}

\begin{figure*}[t]
\centering
\includegraphics[width=0.98\textwidth]{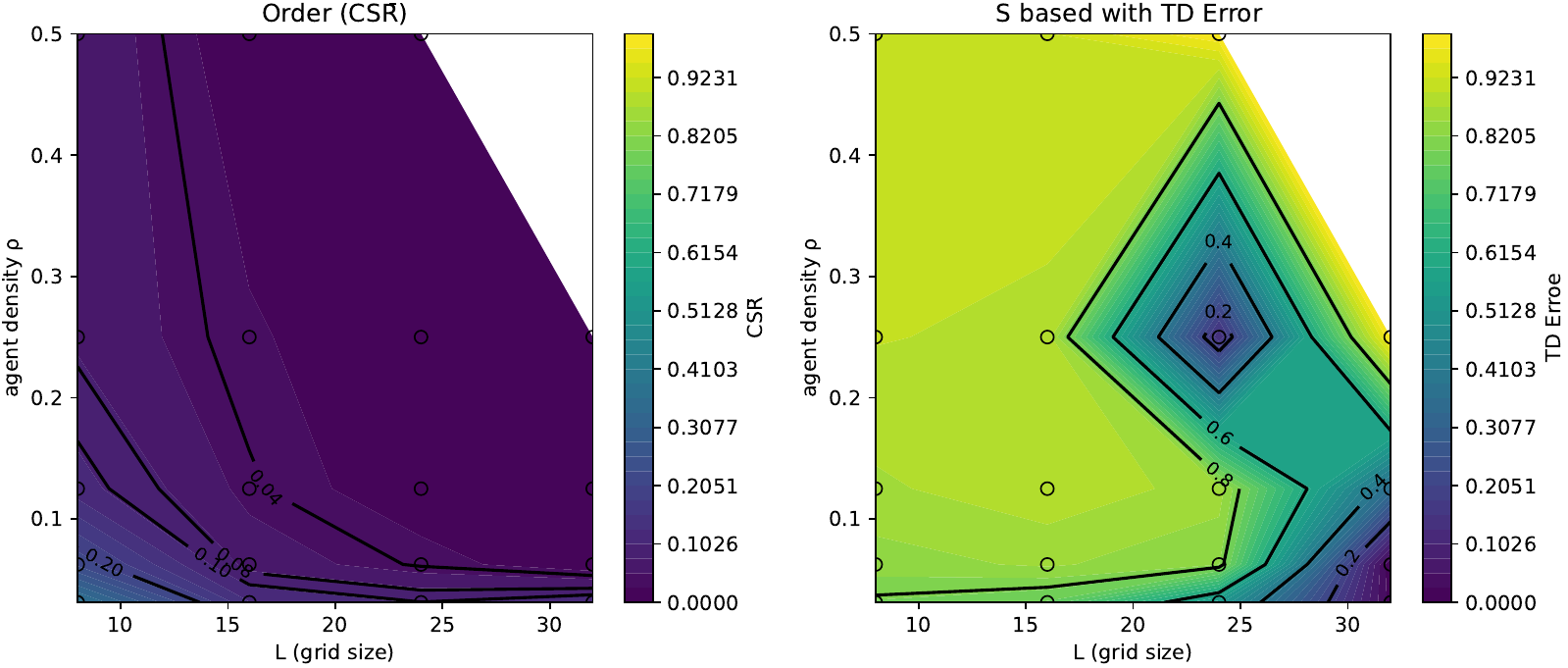}
\caption{
\label{fig:Gv_order}
Phase map using the gradient-norm–based stability index $S_{\nabla}$.
Left: CSR. Right: $S_{\nabla}$.
Global trends resemble the TD-based map, but ridge boundaries become noticeably
less distinct due to optimizer- and curvature-induced fluctuations.
}
\end{figure*}

\begin{figure*}[t]
\centering
\includegraphics[width=0.98\textwidth]{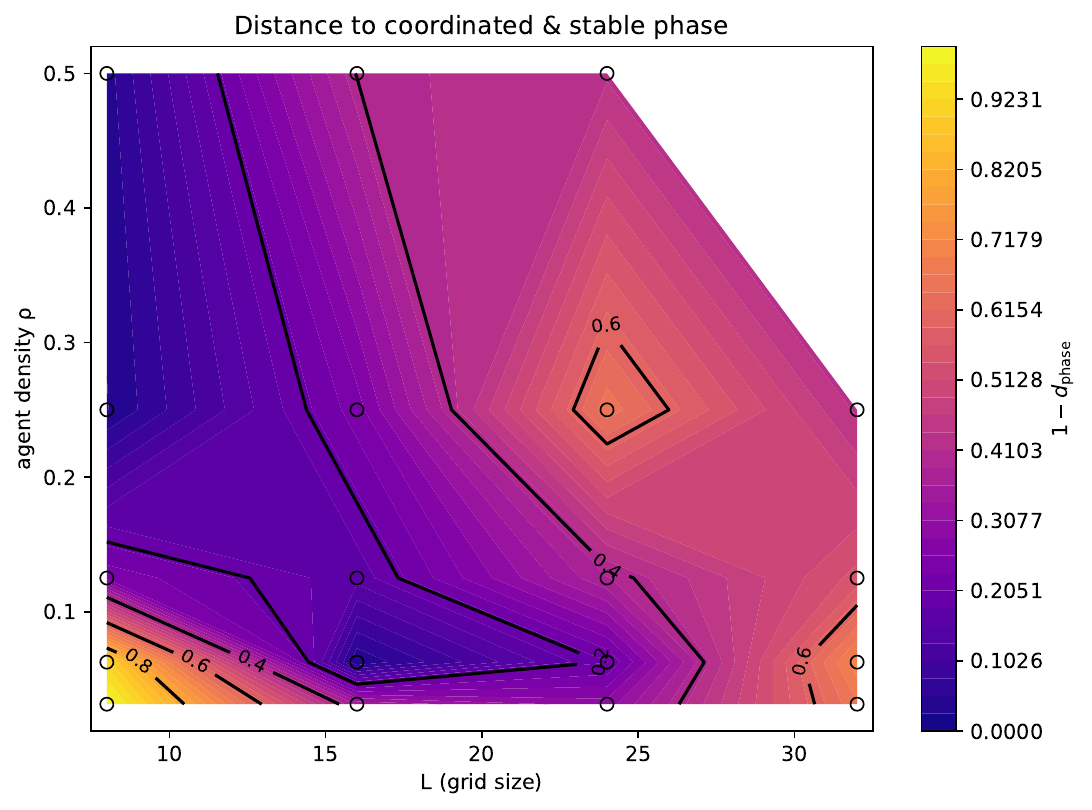}
\caption{
\label{fig:Gvdistance}
Phase-distance map constructed from the gradient-norm–based stability index.
A ridge structure remains visible, but is substantially blurred compared to the
TD-based counterpart.
}
\end{figure*}

In addition to the TD-error–variance stability index $S_{\mathrm{TD}}$, we also
construct a gradient-norm–based stability index $S_{\nabla}$ and visualize the
corresponding phase maps in Figs.~\ref{fig:Gv_order} and \ref{fig:Gvdistance}.

\begin{figure*}[t]
\centering
\begin{subfigure}{0.48\textwidth}
\includegraphics[width=\linewidth]{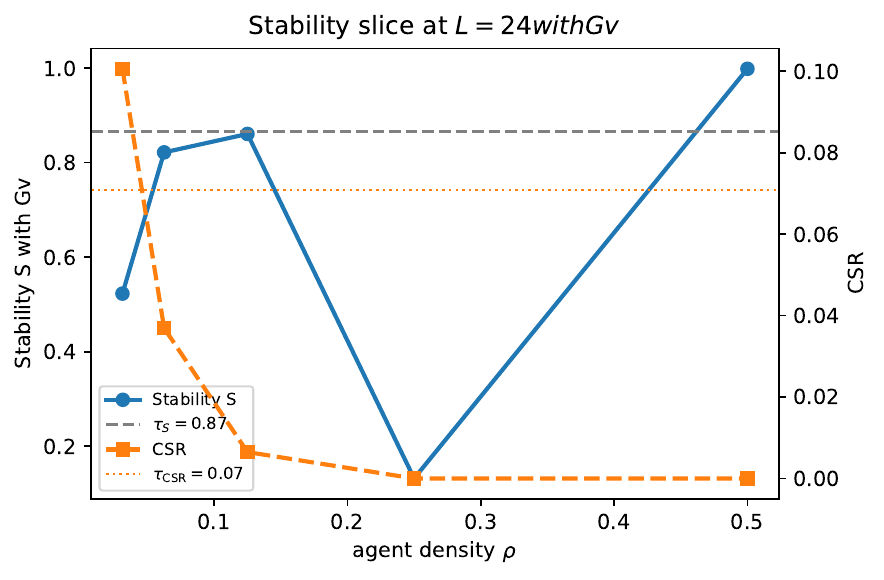}
\caption{$S_{\nabla}$ at $L=24$.}
\label{fig:Gv_slice_L24}
\end{subfigure}
\hfill
\begin{subfigure}{0.48\textwidth}
\includegraphics[width=\linewidth]{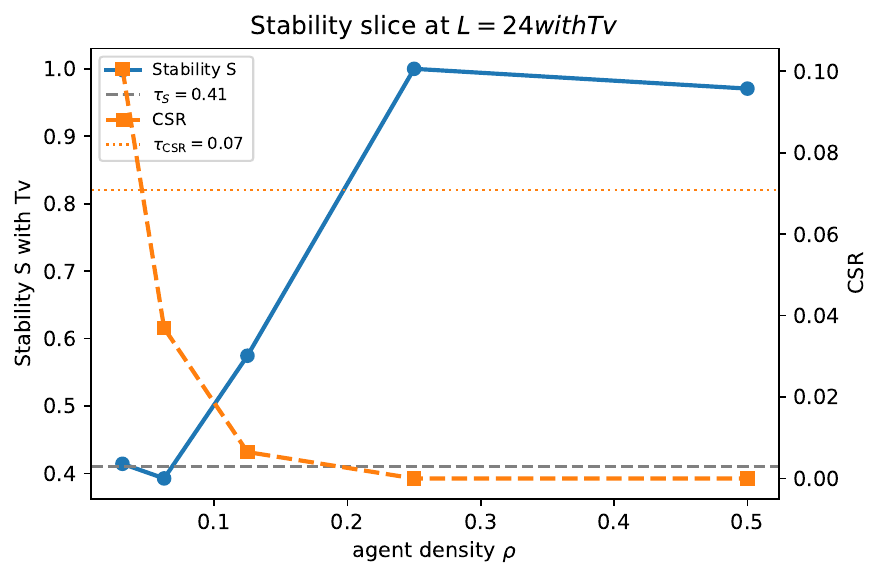}
\caption{$S_{\mathrm{TD}}$ at $L=24$.}
\label{fig:Tv_slice_L24}
\end{subfigure}
\caption{
\label{fig:stability_slice}
Stability-index slices at $L=24$.
Left: $S_{\nabla}$ shows a local collapse at $\rho=0.25$, which obscures the
phase boundary.
Right: $S_{\mathrm{TD}}$ exhibits a distinct peak near $\rho \approx 0.06$--$0.08$,
sharply locating the Instability Ridge.
}
\end{figure*}

At the global level, $S_{\nabla}$ recovers the same broad structure as
$S_{\mathrm{TD}}$: a coordinated and stable phase at small scale and low density,
and a jammed/disordered phase at high density.  
However, the phase boundaries become noticeably less sharp under $S_{\nabla}$.
In particular, at $L=24$ and $\rho=0.25$, $S_{\nabla}$ exhibits a pronounced
local drop (Fig.~\ref{fig:Gv_slice_L24}), whereas $S_{\mathrm{TD}}$ shows a clear
peak around $\rho \approx 0.06$--$0.08$ (Fig.~\ref{fig:Tv_slice_L24}),
precisely identifying the Instability Ridge.

This discrepancy arises because gradient-norm variance is sensitive not only to
kernel drift---the source of distributional non-stationarity---but also to
curvature of the local loss landscape and optimizer momentum.
For instance, at $(L,\rho)=(24,0.25)$, CSR and arrival-time spread remain in a
metastable semi-jammed configuration, yet gradient norms fluctuate due to
fine-scale pattern adjustments, which blurs the phase boundary at the map level.

Taken together, these observations indicate that $S_{\mathrm{TD}}$ is the more
reliable primary indicator for detecting the growth of kernel drift and for
identifying the Instability Ridge.  
We therefore present the $S_{\nabla}$–based maps only as supplementary analysis.

\subsection{Kernel Drift Dynamics Across Scales}\label{seq:Learning_stability_WITHID}

Fig.~\ref{fig:learning_stability_withID} reports the TD error, gradient norm, and their variances across all $(L,\rho)$ conditions (50 seeds; 95
The main trends are as follows.
\begin{enumerate}
\item  TD-error mean
Across all settings the mean TD error quickly approaches zero, indicating that the Bellman update itself stabilizes early.

\item TD-error variance
$\mathrm{Var}[\mathrm{TD}]$ depends primarily on scale $L$, not density.
For small grids ($L=8,16$) the variance is larger and more dynamic, whereas for large grids ($L=24,32$) it remains small and often nearly flat—especially in jammed/disordered regimes such as $(L=24,\rho>0.125)$ and $(L=32,\rho>0.0625)$.

\item  Gradient norm and its variance
Gradient statistics also exhibit strong scale dependence.
Large grids show late-episode divergence at specific densities (e.g., $L=24,\rho=0.25$ and $L=32,\rho=0.125$), indicating gradient-noise–dominated instability even when drift is weak.

\item Density effects
Density modulates the timing and magnitude of fluctuations but does not produce a consistent monotonic trend across scales.
Scale $L$ and drift–synchronization competition near the Instability Ridge (discussed in Fig.~4) dominate the global behavior.

\item Overall structure
These patterns complement the phase diagram in Fig.~\ref{fig:distance}, showing that:
\begin{itemize}
\item drift suppression yields stable coordination,
\item drift–synchronization competition produces fragile dynamics, and
\item weak drift with strong gradient noise leads to jammed/disordered outcomes.
\end{itemize}
\end{enumerate}

\subsection{Ablation: Removing Agent ID and Symmetry Breaking}
\label{seq:app_noid}

To isolate the role of symmetry breaking in the emergence of kernel drift and the
phase structure reported in the main text, we conduct ablations in which agent
identifiers (IDs) are removed. Without IDs, all agents share identical observation
and policy spaces; hence their policy mappings become fully symmetric.

Following the definition of the effective transition kernel in Eq.~\ref{Eq:KL},
removing agent IDs makes all agents share identical observation and policy spaces.
As a result, each agent $i$ satisfies
\[
P_i^t(s'|s,a_i) \approx \bar P, \qquad
\Delta P_i^t := P_i^t - \bar P \approx 0,
\]
and the asymmetry-driven component $\Delta P_{\mathrm{brk}}^t$ in the decomposition
$\Delta P_i^t = \Delta P_{\mathrm{sym}}^t + \Delta P_{\mathrm{brk}}^t$ vanishes.
Kernel drift therefore cannot accumulate.

\paragraph{Learning dynamics without IDs.}
The TD-error mean decreases rapidly at the beginning, similar to the ID-present
case, but then exhibits a secondary rise followed by a slow, noise-like decay.
The TD-error variance shows an early peak comparable to the ID-present setting,
yet does not settle; instead, it undergoes repeated small-amplitude fluctuations.
These patterns indicate that distributional drift is suppressed, and that the
remaining non-stationarity is dominated by weak exploration noise arising from
policy symmetry.

\paragraph{Gradient-norm statistics.}
Removing IDs produces a sharper distinction: gradient magnitudes become
substantially larger---often several times those observed with IDs---and the
corresponding variance rises steeply before decaying slowly. When all agents
share an identical policy, update directions lose diversity, so even small
perturbations are coherently amplified, creating persistent and exaggerated
update noise.

\paragraph{Consequences for phase behavior.}
Overall, removing IDs yields a characteristic structure in which
(i) TD-error variance remains suppressed, confirming the disappearance of
kernel drift, while
(ii) enlarged gradient norms and heightened gradient-norm variance show that
update noise becomes the dominant driver of instability.
This produces superficially similar ``non-coordination'' outcomes to the
jammed/disordered regime observed with IDs, but the underlying mechanism is
fundamentally different: the instability arises not from drift, but from
homogeneous, noise-amplifying updates due to complete policy symmetry.

These findings demonstrate that the phase transitions in the main text are driven by
\[
\text{kernel drift} \ \propto \ \Delta P_{\mathrm{brk}}^t,
\]
i.e., by small but persistent inter-agent asymmetries. Removing such asymmetries
suppresses kernel drift and eliminates both coordinated and fragile regimes.
This result is consistent with prior work on symmetry breaking in learning
dynamics~\cite{Tuyls2005,Bloembergen2015}, role emergence in MARL~\cite{baker2020},
and self-organized differentiation in collective robotic systems~\cite{Bratsun2024}.
It supports the interpretation that the coordination transition observed in
decentralized MARL arises as a spontaneous distribution--interaction phenomenon
rather than an architectural artifact.

This appendix is self-contained and provides supplementary analyses and
ablations that support and extend the results presented in the main text.

\section{Appendix Figures}
\label{sec:appendix_figures}

\subsection{Phase Maps Based on Gradient-Norm Variance}
\label{sec:app_fig_phase_gv}

\subsection{Stability Index Slices at Fixed Scale}
\label{sec:app_fig_stability_slices}

\subsection{Synchronization and Coordination Dynamics}
\label{sec:app_fig_sync_dyn}

\begin{figure}[t]
  \centering
  \begin{subfigure}[t]{0.48\linewidth}
    \centering
    \includegraphics[width=\linewidth]{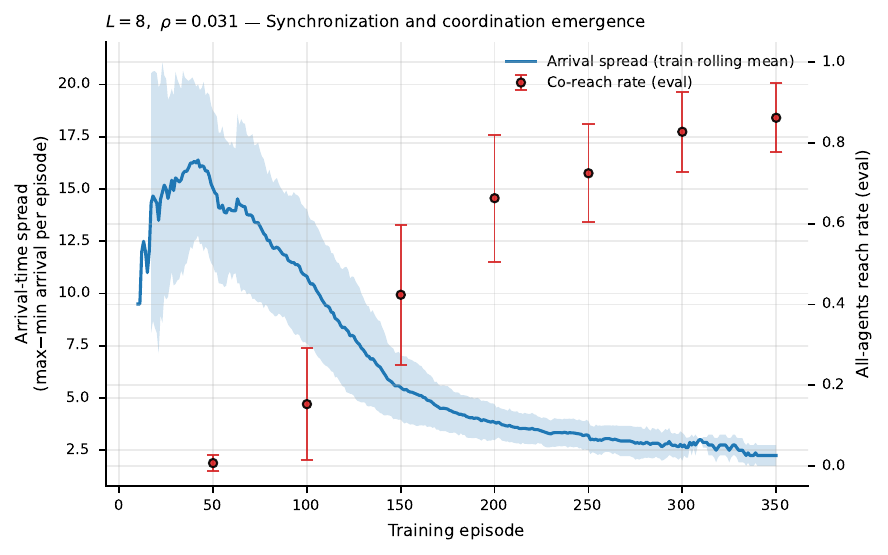}
    \caption{$L=8,\ \rho=0.03125$}
    \label{fig:synccoord_L8_rho0031}
  \end{subfigure}
  \hfill
  \begin{subfigure}[t]{0.48\linewidth}
    \centering
    \includegraphics[width=\linewidth]{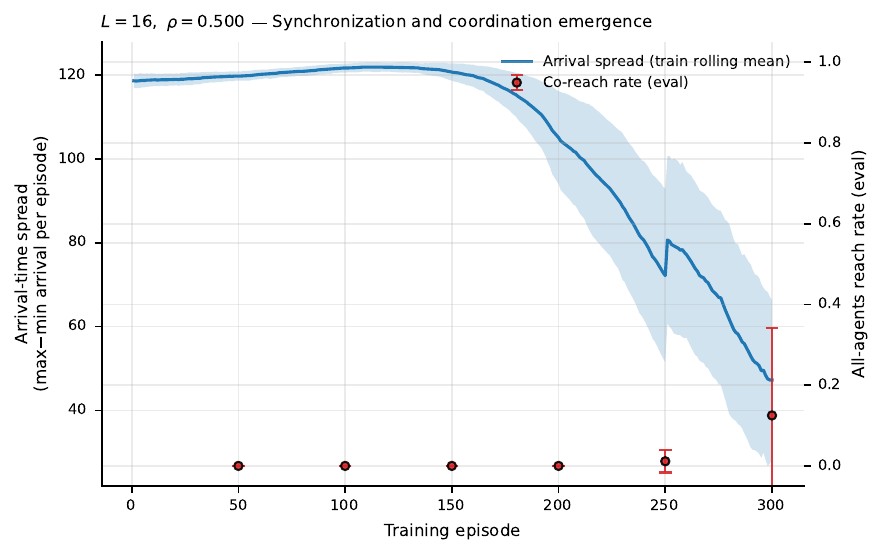}
    \caption{$L=16,\ \rho=0.50$}
    \label{fig:synccoord_L16_rho050}
  \end{subfigure}

  \vspace{0.5em}

  \begin{subfigure}[t]{0.48\linewidth}
    \centering
    \includegraphics[width=\linewidth]{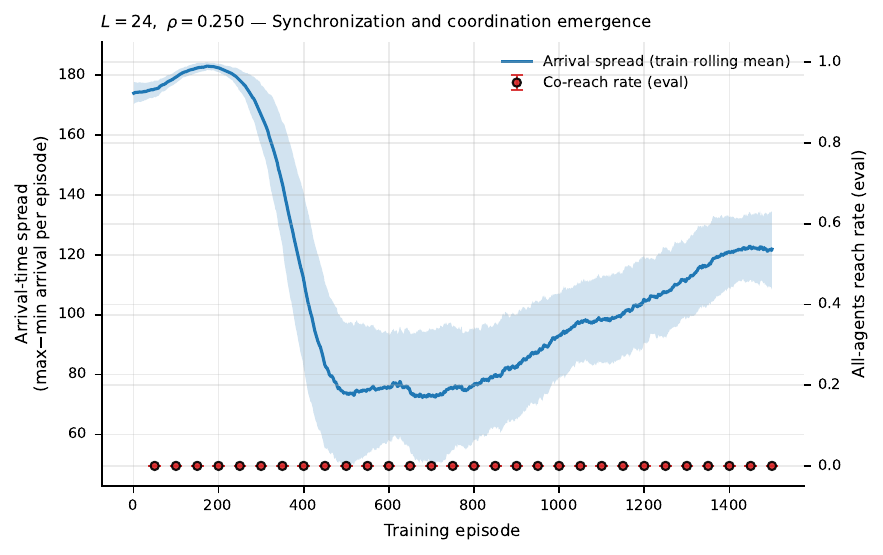}
    \caption{$L=24,\ \rho=0.25$}
    \label{fig:synccoord_L24_rho025}
  \end{subfigure}
  \hfill
  \begin{subfigure}[t]{0.48\linewidth}
    \centering
    \includegraphics[width=\linewidth]{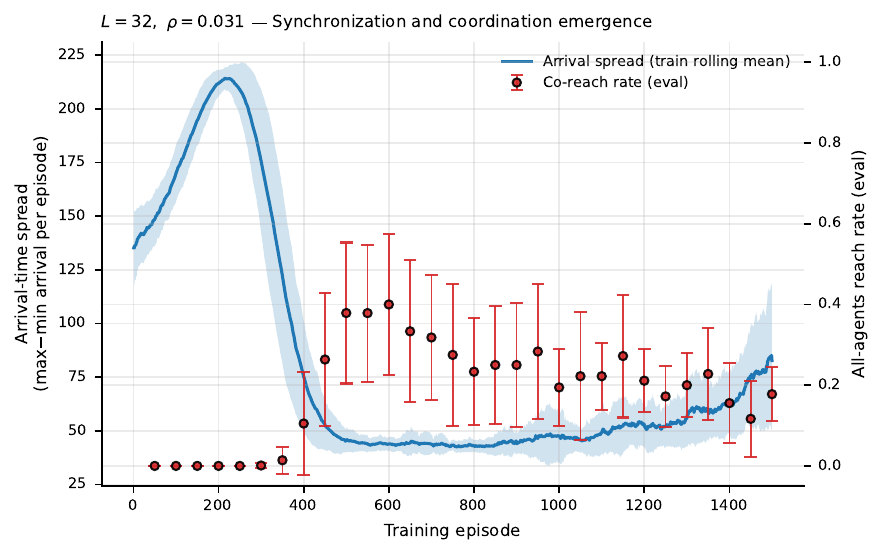}
    \caption{$L=32,\ \rho=0.03125$}
    \label{fig:synccoord_L32_rho0031}
  \end{subfigure}

  \caption{
   Time-series dynamics of arrival-time spread (synchronization) and co-reach rate (coordination)
across four representative conditions.
(a) $L=8,\ \rho=0.03125$: rapid synchronization and stable coordinated behavior.
(b) $L=16,\ \rho=0.50$: approaching the Instability Ridge from the high-density side, 
where synchronization is hindered and co-reach exhibits large fluctuations.
(c) $L=24,\ \rho=0.25$: a jammed/disordered regime outside the Ridge, with persistently large spread and low co-reach.
(d) $L=32,\ \rho=0.03125$: approaching the Ridge from the low-density, large-scale side, 
showing partial synchronization with drift-induced fluctuations.
  }
  \label{fig:synccoord_2x2}
\end{figure}

\subsection{Kernel Drift and Gradient-Noise Regimes}
\label{sec:app_fig_kerneldrift}

\begin{figure}[t]
\centering
\includegraphics[width=\columnwidth]{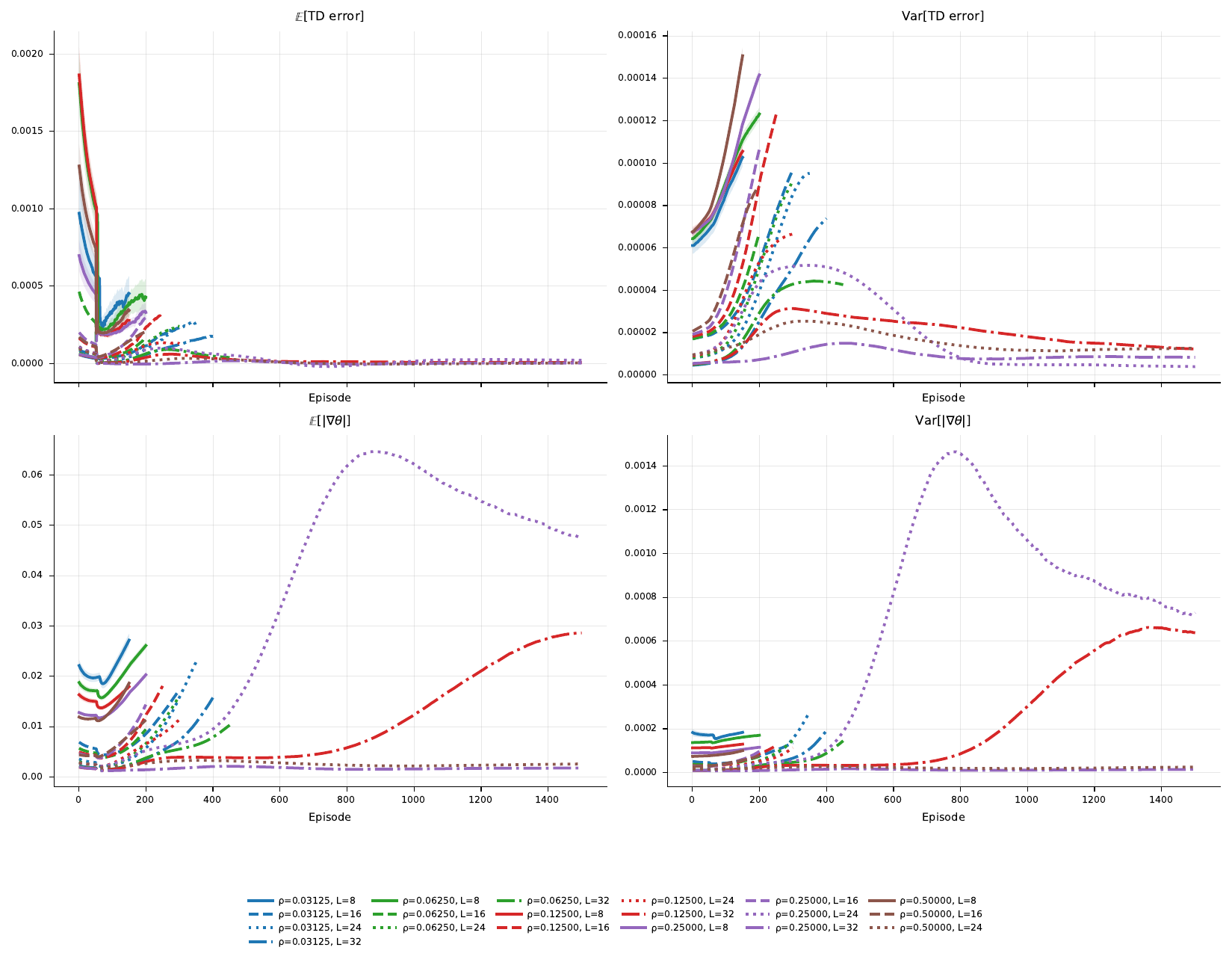}
\caption{TD-error and gradient-norm statistics across all $(L,\rho)$ conditions.
Both metrics exhibit strong scale dependence: larger grids ($L=24,32$) suppress 
TD-error variance except near the Instability Ridge, whereas gradient-norm variance 
shows late-episode divergence only in jammed/disordered regimes 
(e.g., $L=24,\rho=0.25$; $L=32,\rho=0.125$).  
These patterns reflect a transition from drift-dominated to gradient-noise–dominated 
instability. Averaged over 50 seeds; 95\% CIs shown.
}
\label{fig:learning_stability_withID}
\end{figure}

----------------------------------------------------------
\subsection{Effect of Removing Agent IDs}
\label{sec:app_fig_noid}

\begin{figure}[t]
\centering
\includegraphics[width=\columnwidth]{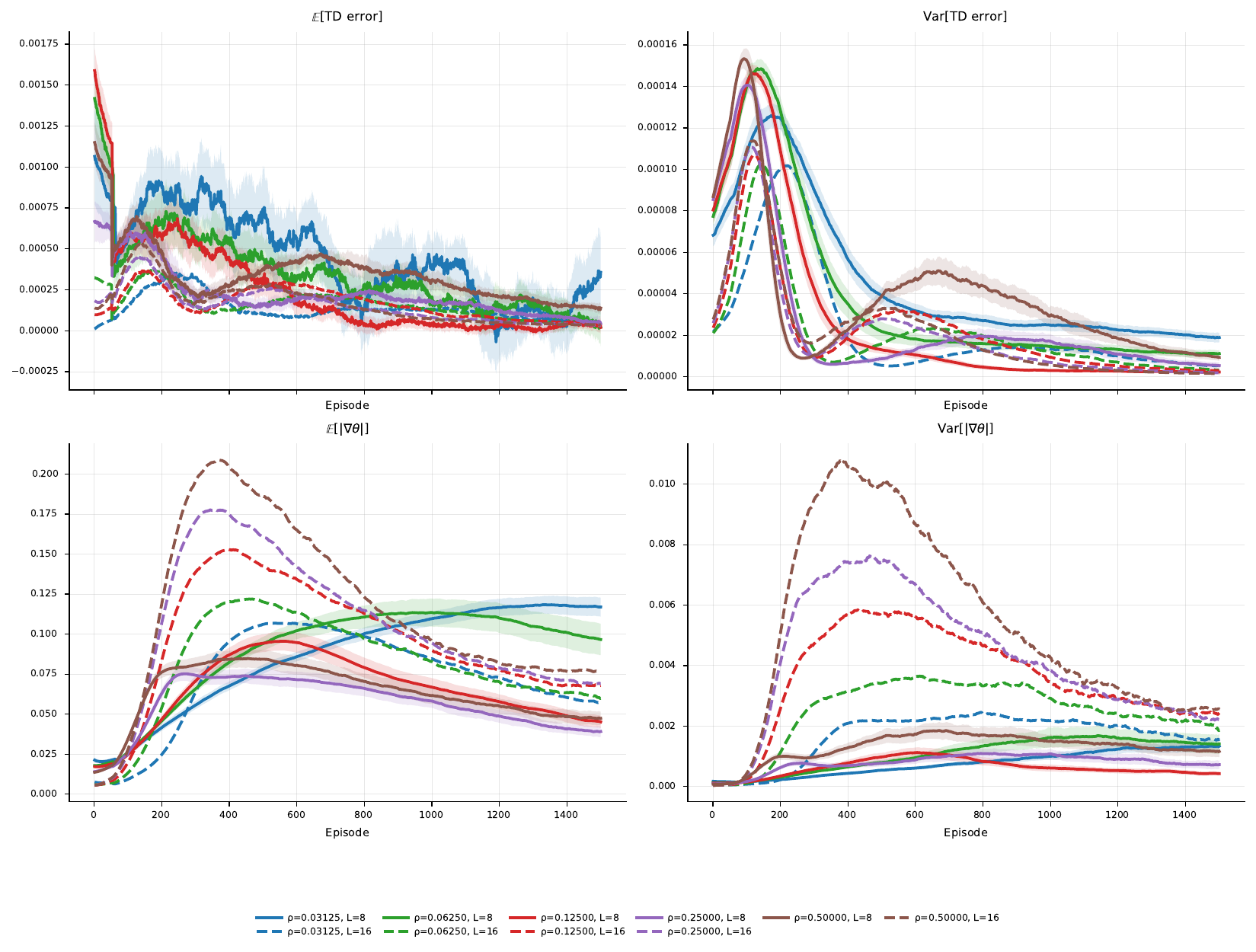}
\caption{
Mean and variance of the TD error (top) and gradient norm (bottom) for all 
$(L,\rho)$ conditions in the ID-removed setting. 
Removing agent IDs suppresses the asymmetry-driven component of kernel drift, 
resulting in low and non-accumulating TD-error variance. 
In contrast, gradient norms remain substantially larger and exhibit pronounced 
early-episode fluctuations, reflecting update-noise amplification under complete 
policy symmetry. 
Curves show averages over 25 seeds with 95\% confidence intervals 
for $L\in\{8,16\}$ and $\rho\in\{0.03125,\dots,0.5\}$.
}
\label{fig:appendix_learning_panel}
\end{figure}

\begin{figure}[t]
\centering
\begin{subfigure}{0.48\textwidth}
    \centering
    \includegraphics[width=\linewidth]{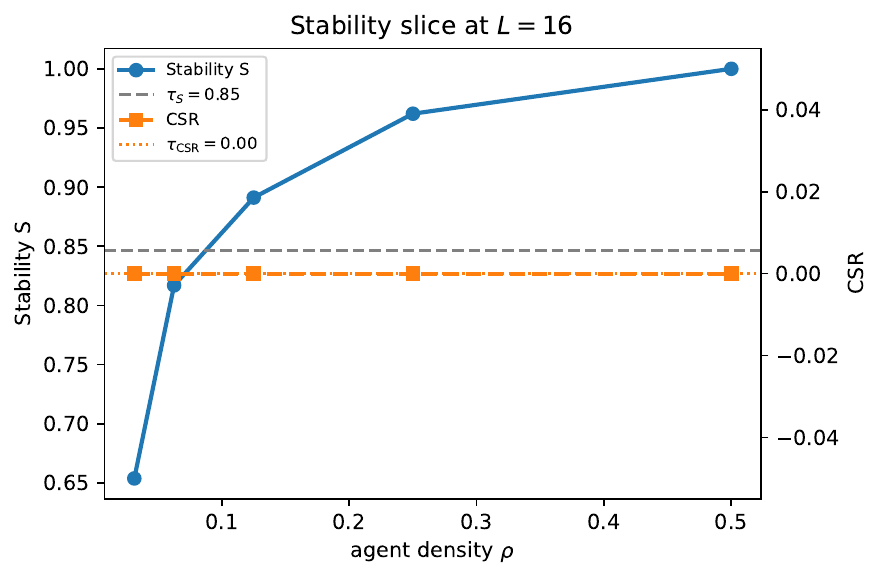}
    \label{fig:CSR_TvS_slice_L16}
\end{subfigure}
\hfill
\begin{subfigure}{0.48\textwidth}
    \centering
    \includegraphics[width=\linewidth]{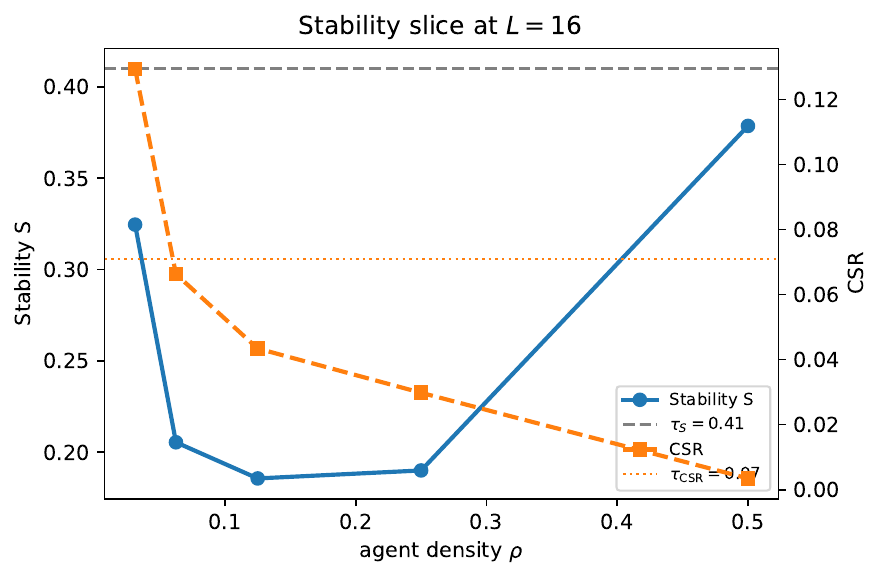}
    \label{fig:CSR_TvS_slice_L16_NOID}
\end{subfigure}
\caption{
Left: \textbf{Without agent IDs}---CSR and the stability index $S$ remain flat across 
densities; no coordinated, fragile, or jammed phases emerge.
Right: \textbf{With IDs}---the coordinated, fragile, and jammed regimes re-emerge, 
confirming that symmetry breaking is required for sustaining kernel drift and for 
producing the phase structure observed in the main text.
}
\label{fig:CSR_S_slice}
\end{figure}

\end{document}